\documentclass[10pt,journal,cspaper]{IEEEtran}

\usepackage{amsmath}
\usepackage{amssymb}
\usepackage{algorithmic}
\usepackage{algorithm}
\usepackage{array}
\usepackage{tabularx}
\usepackage{multirow}

\newcommand{\bfx}{{\textbf{x}}}

\newcommand{\bfw}{{\textbf{w}}}

\newcommand{\bfbeta}{{\boldsymbol{\beta}}}

\newcommand{\bfgamma}{{\boldsymbol{\gamma}}}
\newcommand{\bfdelta}{{\boldsymbol{\delta}}}

\begin{document}

\title{Multi-view learning for multivariate performance measures optimization}

\author{Jim Jing-Yan Wang}

\IEEEcompsoctitleabstractindextext{%
\begin{abstract}
In this paper, we propose the problem of optimizing multivariate performance measures from multi-view data, and an effective method to solve it. This problem has two features: the data points are presented by multiple views, and the target of learning is to optimize complex multivariate performance measures. We propose to learn a linear discriminant functions for each view, and combine them to construct a overall multivariate mapping function for mult-view data. To learn the parameters of the linear discriminant functions of different views to optimize multivariate performance measures, we formulate a optimization problem. In this problem, we propose to minimize the complexity of the linear discriminant functions of each view, encourage the consistences of the responses of different views over the same data points, and minimize the upper boundary of a given multivariate performance measure. To optimize this problem, we employ the cutting-plane method in an iterative algorithm. In each iteration, we update a set of constrains, and optimize the mapping function parameter of each view one by one.
\end{abstract}

}

\maketitle

\IEEEdisplaynotcompsoctitleabstractindextext

\IEEEpeerreviewmaketitle

\section{Introduction}

\IEEEPARstart{D}{ifferent} multivariate performance measures are used to evaluate different machine learning applications \cite{joachims2005support}. For example, in problems of text classification, F1-score and precision/recall
breakeven point (PRBEP) are used to compare true class labels against predicted class labels of a given test text set.
In image retrieval problems, the precision/recall at the top $k$ returned images are used to evaluate the performance of a retrieval system.
Recently, the problem of multivariate performance measures optimization are proposed to learn directly to losses based on pre-defined multivariate performance measures. Many algorithms have been proposed to solve this problem, and some examples are listed as follows.

\begin{itemize}
\item Joachims \cite{joachims2005support} proposed a support vector machine (SVM)-based for multivariate nonlinear performance measures optimization problems.  The proposed algorithm can train a multivariate SVM in polynomial time for large classes of potentially non-linear performance measures. Moreover, the traditional SAM can be arised as a special case of this method.

\item Li et al. \cite{li2013efficient} proposed a two-step approach to optimize multivariate performance measures, by first training a nonlinear classifiers with existing learning methods, and then adapting it to optimize specific performance measures. In the seconde step, the classifier adaptation problem can be solved as a
    quadratic program problem, in a similar way to linear SVM.

\item Mao and Tsang \cite{mao2013feature} proposed a novel feature selection method to optimize multivariate performance measures, by formulating the problem or high-dimensional data, and employing a two-layer cutting plane algorithm to solve it. Moreover, this method is also used in multiple-instance learning problems.
\end{itemize}

Up to now, all these multivariate methods are limited to learn from data with single view. However, in many real-world machine learning problems, the data can be presented by multiple views. For example, in computer vision problems, we can extract different types of features, such as color features, texture features, and shape features, and each feature can be treated as a view. In scientific article classification problems, we can also learn from the views of article abstract, content, and references. Different views of data may be complemental to each other in a learning problem and using multiple views has been a popular strategy in machine learning community.
To learn from multiple views of data, different multi-view learning methods have been proposed \cite{sindhwani2008rkhs,li2002statistical,christoudias2012multi,li2012co}.  However, none of them are designed to optimized a specific multivariate performance measure.

To overcome this problem, in this paper, we propose the problem of learning from multiple view data to optimize the multivariate performance measures. Given a tuple of data points, each of them are presented with multiple views. The problem is to learn a multivariate mapping function to map them to a tuple of class labels, so that the multivariate performance measures can be optimized. To solve this problem, we proposed a novel method for multi-view learning to optimize multivariate performance measures. The contribution of this paper are of two folds:

\begin{enumerate}
\item We propose the problem of multi-view learning for multivariate performance measures. Although there are plenty of multivariate performance measures optimization methods, they are limited to single view data. There are also lots of multi-view learning methods, however, none of them are proposed to optimize multivariate performance measures.

\item We also propose a novel method to solve this problem. We proposed to learn a linear discriminant functions for each view, and combine them to construct a overall multivariate mapping function to predict the class label tuple of a tuple of data points. To learn the linear discriminant functions parameters of different views, we formulate a constrained minimization problem. In this problem, we proposed to minimize the complexity of each linear discriminant functions parameter by minimizing its squared $\ell_2$ norm, encourage the consistence among different views by minimizing the squared $\ell_2$ norm distance among different linear discriminant functions response of differen views, and also minimize the losses based on a specific multivariate performance measures. The minimization problem is optimized by a cutting-plane method in an alterative algorithm \cite{kelley1960cutting}. We maintain a active set of constrains, and update it in each iteration by add a most violated class label tuple. We also optimize the multivariate mapping function parameters one by one, and the optimization of a multivariate mapping function parameter can be solve as a quadratic program problem.

\end{enumerate}

The rest parts of this paper are organized as follows: in section \ref{sec:method}, we introduce the proposed method, and in section \ref{sec:conclusion}, we conclude this paper.

\section{Proposed method}
\label{sec:method}

\subsection{Problem formulation}

We assume we have a training data set of $n$ data points, and each data point has $m$ views. The training set is denoted as $\{(\bfx_{i}^j|_{j=1}^m, y_i)\}|_{i=1}^n$, where $\bfx_i^j\in \mathbb{R}^{d_j}$  is the $d_j$-dimensional feature vector of the $j$-th view of the $i$-th data point, and $y_i \in \{+1,-1\}$ is the binary class label of the $i$-th data point. We propose to learn a multivariate mapping function  $\overline{h}$ to map a tuple of $n$ data points of $m$ views, $\overline{\bfx}=  (\bfx_1^j|_{j=1}^m, \cdots, \bfx_n^j|_{j=1}^m)$, to tuple of $n$ class labels, $\overline{y}=(y_1,\cdots,y_n)$. To implement this
multivariate mapping function, we use a linear discriminant function $f_j(\overline{\bfx}^j,\overline{y}')$ to predict the response of a view the the data tuple, $\overline{\bfx}^j = (\bfx_1^j,\cdots,\bfx_n^j)$, against a candidate class label tuple $\overline{y}' =(y_1',\cdots,y_n')$,

\begin{equation}
\begin{aligned}
f_j(\overline{\bfx}^j,\overline{y}') = \sum_{i=1}^n
\bfw_j^\top \Psi(\overline{\bfx}^j,\overline{y}')
\end{aligned}
\end{equation}
where $\bfw_j\in \mathbb{R}^{d_j}$ is a parameter vector for the linear discriminant function of the $j$-th view, , and $\Psi(\overline{\bfx}^j,\overline{y})$ is a function which can returns a vector to describe the match between $\overline{\bfx}^j$ and $\overline{y}'$, defined as follows,

\begin{equation}
\begin{aligned}
\Psi(\overline{\bfx}^j,\overline{y}')
=
\sum_{i=1}^n {y}_i'\bfx_i^j.
\end{aligned}
\end{equation}
Based on the linear discriminant functions of different views, we construct the multivariate mapping function as follows,

\begin{equation}
\begin{aligned}
\overline{h}(\overline{\bfx})= {\arg\max}_{\overline{y}'\in \mathcal{Y}} \left \{
\sum_{j=1}^m f_j(\overline{\bfx}^j,\overline{y}')=
\sum_{j=1}^m \bfw_j^\top \Psi(\overline{\bfx}^j,\overline{y}')
\right \}
\end{aligned}
\end{equation}
where $\mathcal{Y} = \{+1,-1\}^n$ is a set of all admissible label vectors.

We propose to optimize a complex multivariate performance measure, $\Delta$, by learning the parameter vectors $\bfw_j|_{j=1}^m$ for the $m$ views. To this end, we consider the following three problems,

\begin{enumerate}
\item \textbf{Reducing the complexity of each linear discriminative function}: To prevent over-fitting, we propose to reduce the complexity of the linear discriminative function of each view by minimizing the squared $\ell_2$ norm of its parameter,

\begin{equation}
\begin{aligned}
\min_{\bfw_j|_{j=1}^m}
\frac{1}{2} \sum_{j=1}^m\|\bfw_j\|_2^2.
\end{aligned}
\end{equation}

\item \textbf{Encouraging consistences among different views}: To encourage the consistences of different views, we proposed to minimize the squared  $\ell_2$ norm distances of responses of any two linear discriminative functions  over one single data point,

\begin{equation}
\begin{aligned}
\min_{\bfw_j|_{j=1}^m}
\frac{1}{2}\sum_{j,j':j'<j} \left ( \sum_{i=1}^n \| \bfw_j^\top \bfx_i^j - \bfw_{j'}^\top\bfx_i^{j'} \|_2^2 \right ).
\end{aligned}
\end{equation}

\item \textbf{Optimizing multivariate performance measures}:  To optimize a specific multivariate performance measure, we propose to minimize the upper boundary of a loss function $\Delta$ based on this multivariate performance measure,

\begin{equation}
\begin{aligned}
\min_{\bfw_j|_{j=1}^m, \xi} \xi\\
s.t.
& \forall \overline{y}' \in \mathcal{Y}/\overline{y}:\\
&
\sum_{j=1}^m \bfw_j^\top \left [	  \Psi(\overline{x}^j, \overline{y}) -  \Psi(\overline{x}^j, \overline{y}')\right ]\geq
\Delta(\overline{y}',\overline{y})- \xi,
\end{aligned}
\end{equation}
where $\xi$ is a slack variable of the upper boundary of the loss function.

\end{enumerate}

The overall optimization function are obtained by combining all the problems above,

\begin{equation}
\begin{aligned}
\min_{\bfw_j|_{j=1}^m, \xi}
&\left \{
\frac{1}{2} \sum_{j=1}^m\|\bfw_j\|_2^2 + C_1 \xi \right .\\
&\left .+ \frac{C_2}{2}\sum_{j,j':j'<j} \left ( \sum_{i=1}^n \| \bfw_j^\top \bfx_i^j - \bfw_{j'}^\top\bfx_i^{j'} \|_2^2 \right )
\right \}\\
s.t.
& \forall \overline{y}' \in \mathcal{Y}/\overline{y}:\\
&
\sum_{j=1}^m \bfw_j^\top \left [	  \Psi(\overline{x}^j, \overline{y}) -  \Psi(\overline{x}^j, \overline{y}')\right ]\geq
\Delta(\overline{y}',\overline{y})- \xi.
\end{aligned}
\end{equation}
In the objective of this problem, the first term is to reduce the complexity of the parameter vector of each view, and second term is a slack variable to represent the upper boundary of the multivariate performance measure, and the third term is to encourage the consistences of the responses of different views. $C_1$ and $C_2$ are tradeoff parameters.

\subsection{Problem optimization}

To solve this problem, we employ the cutting-plane algorithm. In an iterative algorithm, we update the parameter vectors $\bfw_j|_{j=1}^m$ and an active set of constrains $\mathcal{W}$ alternately.

\subsubsection{Updating $\bfw_j|_{j=1}^m$} When we have a given active set of constrain $\mathcal{W} \subseteq \mathcal{Y}/\overline{y}$, and optimize the parameter vectors, we optimize $\bfw_j$ one by one, i.e., when $\bfw_j$ is optimized, $\bfw_{j'}|_{j'\neq j}$ is fixed. The following optimization is obtained in this case,

\begin{equation}
\begin{aligned}
\min_{\bfw_j, \xi}
&\left \{
\frac{1}{2} \|\bfw_j\|_2^2 + C_1 \xi
\vphantom{\sum_{j:v_{ij}=1} }
\right .\\
&\left .+ \frac{C_2}{2} \sum_{j':j'<j} \left ( \sum_{i=1}^n \| \bfw_j^\top \bfx_i^j - \bfw_{j'}^\top\bfx_i^{j'} \|_2^2 \right )
\right \}\\
s.t.
& \forall \overline{y}' \in \mathcal{W}:\\
&
\bfw_j^\top \left [	  \Psi(\overline{x}^j, \overline{y}) -  \Psi(\overline{x}^j, \overline{y}')\right ]\\
&
\geq
\Delta(\overline{y}',\overline{y})
-
\sum_{j':j'\neq j} \bfw_{j'}^\top \left [	  \Psi(\overline{x}^{j'}, \overline{y}) -  \Psi(\overline{x}^{j'}, \overline{y}')\right ]
- \xi.
\end{aligned}
\end{equation}
The Lagrange function of this problem is

\begin{equation}
\begin{aligned}
\mathcal{L}&(\bfw_j, \xi, \alpha_{\overline{y}'}|_{\overline{y}': \overline{y}' \in \mathcal{W}})\\
=&
\frac{1}{2} \|\bfw_j\|_2^2 + C_1 \xi
\\
&+ \frac{C_2}{2} \sum_{j':j'<j} \left ( \sum_{i=1}^n \| \bfw_j^\top \bfx_i^j - \bfw_{j'}^\top\bfx_i^{j'} \|_2^2 \right )
\\
&-\sum_{\overline{y}': \overline{y}' \in \mathcal{W}}
\alpha_{\overline{y}'}
\left (
\bfw_j^\top \left [	  \Psi(\overline{x}^j, \overline{y}) -  \Psi(\overline{x}^j, \overline{y}')\right ]
-
\Delta(\overline{y}',\overline{y})
\vphantom{\sum_{j:v_{ij}=1} }
\right .
\\
&
\left .
+
\sum_{j':j'\neq j} \bfw_{j'}^\top \left [	  \Psi(\overline{x}^{j'}, \overline{y}) -  \Psi(\overline{x}^{j'}, \overline{y}')\right ]
+ \xi
\right )\\
=&\frac{1}{2} \bfw_j^\top \Omega \bfw_j   + C_1 \xi  - \bfw_j^\top \bfbeta
+ \frac{C_2}{2} \sum_{j':j'<j} \sum_{i=1}^n \bfw_{j'}^\top \bfx_i^{j'} {\bfx_i^{j'}}^\top \bfw_{j'}\\
&
-\sum_{\overline{y}': \overline{y}' \in \mathcal{W}}
\alpha_{\overline{y}'} \left ( \bfw_j^\top \bfgamma_{\overline{y}'}
- \delta_{\overline{y}'}
+ \xi \right )
\end{aligned}
\end{equation}
where $\alpha_{\overline{y}'} \geq 0$ is the Lagrange multiplier for the constrain $\bfw_j^\top \left [	  \Psi(\overline{x}^j, \overline{y}) -  \Psi(\overline{x}^j, \overline{y}')\right ]
\geq
\Delta(\overline{y}',\overline{y})
-
\sum_{j':j'\neq j} \bfw_{j'}^\top \left [	  \Psi(\overline{x}^{j'}, \overline{y}) -  \Psi(\overline{x}^{j'}, \overline{y}')\right ]
- \xi$, and

\begin{equation}
\label{equ:Omega}
\begin{aligned}
&\Omega = \left ( I + C_2 \sum_{j':j'<j} \sum_{i=1}^n
\bfx_i^j {\bfx_i^j}^\top \right ),\\
&\bfbeta = C_2 \sum_{j':j'<j} \sum_{i=1}^n
\bfx_i^j {\bfx_i^{j'}}^\top \bfw_{j'},\\
&\bfgamma_{\overline{y}'} =\left [	  \Psi(\overline{x}^j, \overline{y}) -  \Psi(\overline{x}^j, \overline{y}')\right ]\\
&\bfdelta_{\overline{y}'} =\left (\Delta(\overline{y}',\overline{y})
-
\sum_{j':j'\neq j} \bfw_{j'}^\top \left [	  \Psi(\overline{x}^{j'}, \overline{y}) -  \Psi(\overline{x}^{j'}, \overline{y}')\right ]
\right ).
\end{aligned}
\end{equation}
This optimization problem can be transferred to its dual form,
\begin{equation}
\label{equ:lag}
\begin{aligned}
\max_{\alpha_{\overline{y}'}|_{\overline{y}': \overline{y}' \in \mathcal{W}}}&
\min_{\bfw_j, \xi}~
\mathcal{L}(\bfw_j, \xi, \alpha_{\overline{y}'}|_{\overline{y}': \overline{y}' \in \mathcal{W}})\\
s.t.&~
\forall \overline{y}', \overline{y}'\in \mathcal{W} :~
\alpha_{\overline{y}'}\geq 0.
\end{aligned}
\end{equation}

To obtain the optimal points of $\bfw_j$ and $\xi$, we set the gradients of Lagrangian function with respect to $\bfw_j$ and $\xi$ to zeros, and we have,

\begin{equation}
\begin{aligned}
&\frac{\partial \mathcal{L}}{\partial \bfw_j} =
\Omega \bfw_j - \bfbeta - \sum_{\overline{y}': \overline{y}' \in \mathcal{W}} \alpha_{\overline{y}'} \bfgamma_{\overline{y}'}=0\\
&\Rightarrow
\bfw_j
=
\Omega^{-1}\left (\bfbeta +\sum_{\overline{y}': \overline{y}' \in \mathcal{W}} \alpha_{\overline{y}'} \bfgamma_{\overline{y}'}\right ),\\
&\frac{\partial \mathcal{L}}{\partial \xi}
=
C_1 -\sum_{\overline{y}': \overline{y}' \in \mathcal{W}}
\alpha_{\overline{y}'}=0\\
& \Rightarrow
 \sum_{\overline{y}': \overline{y}' \in \mathcal{W}}
\alpha_{\overline{y}'}= C_1 .
\end{aligned}
\end{equation}
By substituting these results back to (\ref{equ:lag}), we
obtain the dual problem as

\begin{equation}
\label{equ:lag1}
\begin{aligned}
\max_{\alpha_{\overline{y}'}|_{\overline{y}': \overline{y}' \in \mathcal{W}}}&
\left \{-
\frac{1}{2}
\left (\bfbeta +\sum_{\overline{y}': \overline{y}' \in \mathcal{W}} \alpha_{\overline{y}'} \bfgamma_{\overline{y}'}\right )^\top \Omega^{-1}
\left (\bfbeta +\sum_{\overline{y}': \overline{y}' \in \mathcal{W}} \alpha_{\overline{y}'} \bfgamma_{\overline{y}'}\right ) \right .\\
&\left .-\sum_{\overline{y}': \overline{y}' \in \mathcal{W}}
\alpha_{\overline{y}'}
\delta_{\overline{y}'}
\vphantom{\sum_{j:v_{ij}=1} }
\right \}
\\
s.t.~&
\forall \overline{y}', \overline{y}'\in \mathcal{W} :~
\alpha_{\overline{y}'}\geq 0,~and~ \sum_{\overline{y}': \overline{y}' \in \mathcal{W}}
\alpha_{\overline{y}'}= C_1.
\end{aligned}
\end{equation}
We can solve this problem as a quadratic program problem.

\subsubsection{Updating $\mathcal{W}$} To update $\mathcal{W}$, we first find the most violated $\overline{y}'$ and then add it the $\mathcal{W}$. $\overline{y}'$ is obtained as

\begin{equation}
\label{seq:y}
\begin{aligned}
\overline{y}' =  {\arg\max}_{\overline{y}''\in \mathcal{Y}/y} &\left \{ \Delta(\overline{y}'',\overline{y})
+ \sum_{j=1}^m \bfw_j^\top \Psi(\overline{x}^{j}, \overline{y}'')
\right .\\
&
\left .
- \sum_{j=1}^m \bfw_j^\top \Psi(\overline{x}^{j}, \overline{y})
\right \}.
\end{aligned}
\end{equation}

\subsubsection{Alterative algorithm}

The proposed iterative algorithm is given in Algorithm \ref{alg}.

\begin{algorithm}[h!]
\caption{Iterative multi-view learning algorithm for multivariate performance measures.}
\label{alg}
\begin{algorithmic}
\STATE \textbf{Input}: Multi-view training data set $\{(\bfx_{i}^j|_{j=1}^m, y_i)\}|_{i=1}^n$;

\STATE \textbf{Input}: Tradeoff parameters $C_1$ and $C_2$;

\STATE \textbf{Input}: The maximum iteration number $T$.

\STATE Initialize linear discriminative function parameter $\bfw_j^0$ for each view;

\STATE $\mathcal{W} \leftarrow \emptyset$;

\STATE Calculate $\Omega$ as in (\ref{equ:Omega});

\FOR{$t=1,\cdots,T$}

\STATE Find the most violated class label tuple ${\overline{y}'}^t$ as in (\ref{seq:y}) by fixing $\bfw_j^{t-1}|_{j=1}^m$;

\STATE Add ${\overline{y}'}^t$ to the constrain active set $\mathcal{W}$, $\mathcal{W} \leftarrow \mathcal{W}\cup
\{{\overline{y}'}^t\}$;

\FOR{$j=1,\cdots,m$}

\STATE Update $\bfw_j^t$ by solving (\ref{equ:lag1}) by fixing $\bfw_{j'}^{t-1}|_{j'\neq j}$ and $\mathcal{W}$;

\ENDFOR

\ENDFOR

\STATE \textbf{Output}: $\bfw_j^T|_{j=1}^m$.

\end{algorithmic}
\end{algorithm}

\section{Conclusion}
\label{sec:conclusion}

In this paper, we propose the problem of multi-view learning for multivariate performance measures, and a novel algorithm to solve it. Multivariate performance measures optimization can obtain a predictor which directly optimize a desired performance measure. However, the existing methods are limited to single view data, and cannot utilize multi-view data. We propose to learn from multi-view data to optimize the multivariate performance measures,  and it has potential in real-world applications. The proposed method can be implemented easily due to its employment of cutting plane method and quadratic program.


\end{document}